\documentclass[10pt]{article} 
\usepackage[accepted]{tmlr}

\usepackage{amsmath,amsfonts,bm}

\def\eqref#1{equation~\ref{#1}}

\def\1{\bm{1}}

\DeclareMathAlphabet{\mathsfit}{\encodingdefault}{\sfdefault}{m}{sl}
\SetMathAlphabet{\mathsfit}{bold}{\encodingdefault}{\sfdefault}{bx}{n}

\usepackage{array}
\usepackage{hyperref}
\usepackage{url}
\usepackage{booktabs}
\usepackage{xcolor}
\usepackage{graphicx}

\def\openreview{}

\title{Continuous Output Personality Detection Models via Mixed \\ Strategy Training}

\author{Rong Wang\thanks{Equal contributions} \\
	NLP Center\\
	University of Stuttgart\\
	\texttt{rongw.de@gmail.com} 
	\AND
	Kun Sun\footnotemark[1] \\
	Tübingen\\
	\texttt{sharpksun@hotmail.com}
}

\def\month{MM}  
\def\year{YYYY} 

\begin{document}
	
	\maketitle

\begin{abstract}
 The traditional personality models only yield binary results.  This paper presents a novel approach for training personality detection models that produce continuous output values, using mixed strategies. By leveraging the PANDORA dataset, which includes extensive personality labeling of Reddit comments, we developed models that predict the Big Five personality traits with high accuracy. Our approach involves fine-tuning a RoBERTa-base model with various strategies such as Multi-Layer Perceptron (MLP) integration, and hyperparameter tuning. The results demonstrate that our models significantly outperform traditional binary classification methods, offering precise continuous outputs for personality traits, thus enhancing applications in AI, psychology, human resources, marketing and health care fields.
\end{abstract}

\section{Introduction}

Personality is a fundamental aspect of human psychology that encompasses the combination of behaviors, emotions, motivations, and thought patterns that define an individual. Understanding personality is crucial as it influences a wide array of life outcomes, including individual well-being, mental and physical health, interpersonal relationships, career choices, and social behaviors. Given its pervasive impact, the ability to accurately assess and predict personality traits has significant implications across various fields, from psychology and education to human resources and marketing. Personality prediction and recognition is a key issue in artificial intelligence (AI), natural language processing (NLP), sentiment analysis, and computational cognition. 

Among the various cognitive theories of human personality, the \texttt{Big Five} or \texttt{Five Factor Model} is one of the most widely accepted frameworks \citep{digman1990personality}. The \texttt{Big Five} theory categorizes personality traits into five broad dimensions: Extraversion (assertive, energetic, outgoing), Agreeableness (appreciative, generous, compassionate), Conscientiousness (efficient, organized, responsible), Neuroticism (anxious, self-pitying, worried), and Openness to Experience (curious, imaginative, open-minded). These dimensions provide a comprehensive framework for understanding and measuring personality.

Traditionally, particularly in psychology research,  personality traits are assessed through standardized questionnaires or surveys, such as the NEO Five Factor Inventory and the Big-Five Inventory \citep{john1991big, fossati2011big}, which require individuals to reflect on their typical patterns of thinking and behavior. Another widely used framework, particularly in applied settings, is the Myers-Briggs Type Indicator (MBTI) \cite{myers1987introduction}. Unlike the Big Five, MBTI conceptualizes personality as a set of continuous traits. Specifically, the MBTI categorizes personality into 16 distinct types based on four binary dimensions: Extraversion/Introversion, Sensing/Intuition, Thinking/Feeling, and Judging/Perceiving. These research methods play a crucial role in theory building and data collection, and based on these frameworks, several datasets have been established. 

Psychologists have long recognized the importance of language in uncovering personality traits. For instance, word usage can reveal emotional states, cognitive processes, and social behaviors. To aid in this, psychologists have developed emotion lexicons that serve as valuable references for assessment and research. Leveraging these lexicons, researchers have applied NLP and machine learning techniques to evaluate personality traits from textual and spoken data \citep{kahn2007measuring, cambria2010senticnet, mohammad2013crowdsourcing}. 

The advancements in NLP and machine learning have created new opportunities for assessing personality through the analysis of verbal behavior, including text and speech. Language serves as a rich medium that reflects an individual's underlying motivations, emotions, and cognitive patterns. NLP and machine learning have been used to develop automated models capable of recognizing and predicting personality traits by analyzing text and speech. For example, some automatic personality prediction models were created to help do such tasks. Early approaches to automatic personality prediction from text relied on machine learning models that used psycholinguistic features—specific language cues linked to psychological states. These models extracted features such as word frequencies, sentence structures, and semantic content to predict personality traits. The machine learning methods used for personality detection mainly include SVM, Bayeisan, Gradient Boosting, etc. \citep{mairesse2007using, amirhosseini2020machine}. 

However, more recent approaches have leveraged deep learning techniques (such as LSTM, CNN etc.) and pre-trained word embeddings, which capture contextual information from large corpora of text \citep{poria2013common, majumder2017deep, mehta2020recent}. Despite of this, the state-of-the-art (SOTA) development is to integrate comprehensive psycholinguistic features with advanced transformer-based language models or large language models (e.g., BERT, BART etc.) to enhance the accuracy and interpretability of personality predictions \citep{mehta2020bottom, christian2021text, ramezani2022automatic, lin2023novel, hilliard2024eliciting}. In this sense, the transformer-based models have been largely applied to automatic personality detection, and these models have achieved the SOTA results.


The SOTA personality detection models need to be trained on appropriate datasets. Experiments are typically conducted on widely used benchmark datasets, such as the Big Five Essay dataset (referred to as ``Essay'')\citep{pennebaker1999linguistic} and the MBTI Kaggle dataset (short as ``MBIT'')\citep{li2018feature}. Past research on personality detection has primarily focused on binary classification (negative or positive) across different dimensions. The training datasets often employed include ``Essay'' or ``MBTI''. However, the size of these training datasets is relatively small; for instance, the ``Essay'' dataset contains only \textbf{2468} texts. More importantly, users expect personality models to produce results with continuous values rather than binary outputs. For example, users could expect that such a result: ``agreeableness: 0.23; openness: 0.47; conscientiousness: 0.39; extraversion: 0.31; neuroticism: 0.78'', rather than ``agreeableness: N; openness: P; conscientiousness: P; extraversion: N; neuroticism: P (N= negative, P =Positive)''. Clearly, training models to produce continuous values as outputs is more challenging than developing models that yield binary results.

To address these challenges, the current study proposes effective methods for training personality models based on existing classifier models using a new training dataset. The newly trained personality models can much more effectively and accurately recognize personality traits with continuous output values for input texts. Moreover, our methods can be applied to train similar classifiers to produce continuous output values.

\section{Methods}

This section introduces the new training dataset for personality assessment and details training strategies designed to enhance the capabilities of such classifiers. The new dataset includes a larger and more diverse collection of texts to address the limitations of previous datasets. Additionally, advanced techniques are employed to improve model performance. These strategies aim to ensure that the classifier not only handles a wider range of inputs but also produces more accurate and reliable continuous output values.

\subsection{Dataset}

The PANDORA dataset is a comprehensive compilation of Reddit comments from over 10,000 users, labeled with three personality models (Big Five, MBTI, Enneagram) and demographic data (age, gender, location) \citep{gjurkovic-etal-2021-pandora}. This dataset is the first to provide such extensive labeling, including Big Five personality traits for 1,600 users, alongside millions of comments that offer rich linguistic data for analysis. By integrating these diverse labels, PANDORA addresses the scarcity of datasets that combine personality and demographic information, offering a valuable resource for computational sociolinguistics and NLP research.

The dataset facilitates a range of experiments and analyses. It enables researchers to predict Big Five traits using more readily available MBTI and Enneagram labels, analyze gender classification biases, and explore relationships between psychodemographic variables and language use. Extracting Big Five labels from \texttt{Reddit} comments involves identifying comments where users reported their Big Five test results. These results were then manually verified and normalized to ensure accuracy. The dataset includes comments from 1608 users labeled with Big Five traits, providing a substantial amount of data for analysis. 

There are \textbf{16048} texts in the training dataset, 2416 evaluation dataset, and 2416 texts in the test dataset. The data on Big Five are continuous types, and they are ideal training data for our desired models. The data on MBTI are still binary, which are excluded in the present study. 

\subsection{Training strategies}
Given that scoring is a multiple classification task, we can fine-tune existing high-performance models to tailor them for producing continuous output values. To validate our strategies, we selected a prominent existing model as the foundational model for fine-tuning. Building upon this model, we implemented additional strategies to enhance its capabilities. By employing these various strategies, we were able to create multiple models for personality detection, each optimized for different aspects of the task. The following details how we implement our training procedures. 

 This RoBERTa-base model \citep{camacho2022tweetnlp} performed quite well in twitter sentiment analysis.  
This RoBERTa-base model is good at making classifications, and we can take advantage of this classifier to fine-tune. The RoBERTa also works well in making fine-tuning.

\begin{figure}[htbp]
    \centering
    \fbox{\includegraphics[width=0.8\textwidth]{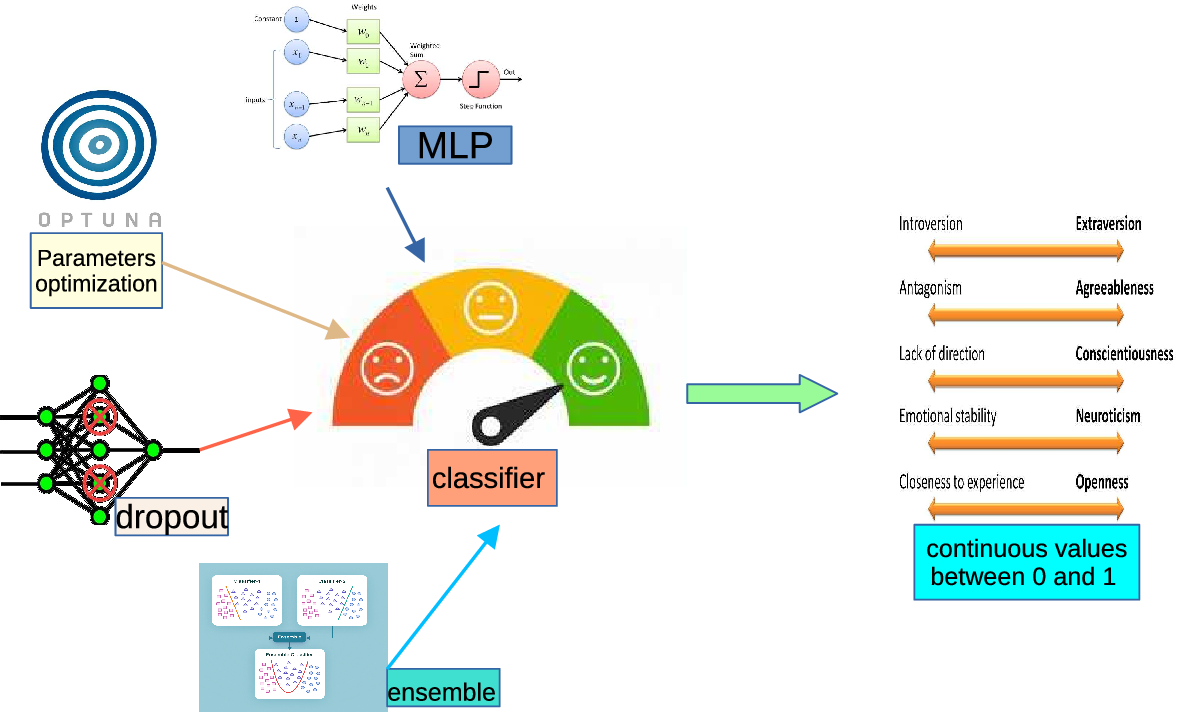}}
    \caption{The roadmap of the present study}
    \label{fig:roadmap1}
\end{figure}

Based on this model, we adopted the strategies shown in Table~\ref{table:str}. We provide a detailed account of these strategies one by one. A Multi-Layer Perceptron (MLP) is a class of feedforward artificial neural networks consisting of at least three layers of nodes: an input layer, one or more hidden layers, and an output layer. We trained the entire model, including both the pre-trained layers and the new MLP layers, on the training dataset to adjust the weights of both the pre-trained model and the new MLP layers. Hyperparameter tuning was extended using \texttt{Optuna} to explore a broader range of hyperparameters, as seen in ``S2-5''. Additional strategies were implemented in ``S4'' and ``S5''. For example, a more sophisticated learning rate scheduler was used, along with regularization techniques such as dropout and L2 regularization to reduce overfitting. Mixed precision training, which combines different numerical precisions, was employed to enable faster and more efficient training on GPUs in ``S3'', ``S4'', and ``S5''. Data augmentation was used in ``S5'' to increase the diversity of training data by augmenting the text. Additionally, ensemble learning, which involves combining the predictions of multiple models to improve overall performance, was adopted in ``S5''. The existing models and training strategies are depicted in Figure~\ref{fig:roadmap1}. The details of the training strategies are summarized in Table~\ref{table:str}.

In order to create a basic baseline model, we used \texttt{BERT} (not RoBERTa) for fine-tuning with the same training dataset. This strategy is labeled as ``S0'' in Table~\ref{table:str}. The aforementioned strategies were chosen to train various models. We created six types of strategies (shown in Table~\ref{table:str}) to train different models. For ``S0'', no special strategies were employed, while ``S1'' used RoBERTa as the pre-trained model for fine-tuning. ``S2'' introduced an MLP and used Optuna to select the best hyperparameters. ``S3'' incorporated mixed precision training. ``S4'' did not include the MLP but retained other setups from ``S3''. ``S5'' essentially used the setups in ``S3'' but added synonym augmentation and ensemble learning. This approach allowed us to create six types of models using different combinations of strategies. \footnote{The trained models are available at: \url{https://huggingface.co/KevSun/Personality_LM}.}

\begin{table}[h!]
\centering
\caption{Summary of Strategies and Details}
\vspace{0.1cm}
\begin{tabular}{|c|>{\raggedright\arraybackslash}p{4cm}|>{\raggedright\arraybackslash}p{8cm}|}
\hline
\textbf{Item} & \textbf{Strategies} & \textbf{Detail} \\ \hline
S0 & Direct fine-tuning on BERT & Manually set hyperparameters \\ \hline
S1 & Direct fine-tuning on RoBERTa & Manually set hyperparameters \\ \hline
S2 & Fine-tuning + select hyperparameters & Optuna for choosing best hyperparameters \\ \hline
S3 & Fine-tuning + MLP + select parameters + Mixed precision training & Add MLP layers in fine-tuning, and Optuna selecting the optimized hyperparameters  \\ \hline
S4 & Fine-tuning + select parameters + Mixed precision training & Basically use the S3 setups but manually change some parameters \\ \hline
S5 & Fine-tuning + MLP + best hyperparameter + SynonymAug + Ensemble learning & Basically follow the setups in S 4 but use synonym to increase the addition of data augmentation \\ \hline
\end{tabular}
\label{table:str}
\end{table}


\subsection{Evaluation metrics}

We used a wide range of metrics to evaluate the model performance. Considering that our models produce continuous output values, we opted for metrics suitable for evaluating continuous values rather than traditional metrics like accuracy and F1 score. The following details these metrics.

\texttt{Mean Squared Error (MSE)} measures the average of the squared differences between predicted and actual values, giving higher weight to larger errors. It is non-negative, with lower values indicating better performance. For instance, an MSE of 0.01 means that, on average, the squared difference between predicted and actual values is 0.01. \texttt{Mean Absolute Error (MAE)} calculates the average of the absolute differences between predicted and actual values and is more robust to outliers compared to MSE. It is also non-negative, with lower values representing better performance. For example, an MAE of 0.1 means that, on average, the absolute difference between predicted and actual values is 0.1. The two metrics were primary ones in the present study.
\texttt{R-squared (R²)} measures the proportion of variance in the dependent variable predictable from the independent variables, ranging from 0 to 1, with higher values indicating better performance. An R² of 0.8 means 80\% of the variance in the dependent variable is predictable from the independent variables. This metric was taken as a supplementary one. 

Meanwhile, we also provided \texttt{accuracy} and \texttt{F1 score} metrics for reference to facilitate comparison with results from past relevant research. A higher accuracy or F1 score indicates better performance. However, accuracy and F1 score are primarily used to evaluate binary classification performance rather than classifiers with continuous output values.

\section{Results}
\label{gen_inst}
Using the same training and evaluation datasets, we trained five different models as outlined in Table~\ref{tab:rest}. Following the training phase, we employed the same test data to evaluate the performance of these models using the evaluation metrics discussed earlier. In addition to overall performance, we assessed the effectiveness of each model in predicting each of the five personality traits individually. This comprehensive evaluation allows for a detailed comparison of how well each model performs across different dimensions of personality. The results, detailing both the overall performance and the performance for each personality dimension, are summarized in Table~\ref{tab:rest}.

\begin{table}[h!]
\centering
\caption{Models' Overall Performance}
\vspace{0.5cm}
\begin{tabular}{lcccccc}
\toprule
\textbf{Model (Strategies)} & \textbf{MSE} & \textbf{MAE} & \textbf{R²} & \textbf{Accuracy} & \textbf{F1 Score} \\
\midrule
\textcolor{red}{M0 (S0)} & \textcolor{red}{0.15} & \textcolor{red}{0.33}& \textcolor{red}{0.22} & \textcolor{red}{0.50} & \textcolor{red}{0.47} \\~\\
M1 (S1) & 0.099 & 0.25 & 0.27 & 0.65 & 0.63 \\~\\
M2 (S2) & 0.10  & 0.25 & 0.26 & 0.64 & 0.62 \\~\\
\textcolor{blue}{M3 (S3)} & \textcolor{blue}{0.07}  & \textcolor{blue}{0.16} & \textcolor{blue}{0.59}  & \textcolor{blue}{0.80} & \textcolor{blue}{0.78} \\~\\
M4 (S4) & 0.1   & 0.25 & 0.26 & 0.64 & 0.60 \\~\\
M5 (S5) & 0.1   & 0.25 & 0.26 & 0.64 & 0.60 \\
\hline\hline
\end{tabular}
\label{tab:rest}
\end{table}

According to Table~\ref{tab:rest}, we have several findings. First, M0 did not demonstrate superior performance compared to the other models. This suggests that models fine-tuned based on BERT do not outperform those fine-tuned based on RoBERTa. It also supports our approach of using RoBERTa for fine-tuning. Second, it appears that M3 has the best performance based on MSE, MAE, $R^2$, accuracy, and F1 score. Specifically, M3 exhibits the smallest MSE and MAE values, coupled with the highest $R^2$, accuracy, and F1 score. These evaluation indices are significantly better than those of the other models. In contrast, the other five models demonstrate similar performance across these metrics, without any one model standing out significantly.

To provide a comprehensive understanding, we detail the performance across the five personality trait dimensions for each model. Here, we specifically highlight the performance of M1 and M3 in each dimension to illustrate the differences and improvements achieved by M3. The remaining models are discussed in the Appendix.

For M1 and M3, we analyze the prediction accuracy for each of the five personality traits: Extraversion, Agreeableness, Conscientiousness, Neuroticism, and Openness to Experience. This detailed examination helps identify which traits are predicted more accurately and to what extent M3 outperforms M1 and other models. Table~\ref{tab:m3} presents a summary of the performance in each dimension in M3. Table~\ref{tab:m1} summarizes the dimension performance.  By focusing on M1 and M3, we can clearly see the enhancements made in M3. This comprehensive evaluation allows us to understand the strengths and limitations of each model in predicting continuous personality traits.

\begin{table}[h!]
\centering
\caption{Model 3(\texttt{M3})'s Performance in Big Five Traits}
\vspace{0.5cm}
\begin{tabular}{lcccccc}
\toprule
\textbf{Big Five} & \textbf{MSE} & \textbf{MAE} & \textbf{R²} & \textbf{Accuracy} & \textbf{F1 Score} \\
\midrule
Agreeableness & 0.05 & 0.58 & 0.41 & 0.79 & 0.78 \\~\\
\textcolor{blue}{Conscientiousness} & \textcolor{blue}{0.03}  & \textcolor{blue}{0.67} & \textcolor{blue}{0.58} & \textcolor{blue}{0.88} & \textcolor{blue}{0.87} \\~\\
Extraversion & 0.21  & 0.35 & 0.31  & 0.61 & 0.47 \\~\\
Neuroticism & 0.20   & 0.31 & 0.32 & 0.62 & 0.48 \\~\\
Openness & 0.03   & 0.12 & 0.66 & 0.84 & 0.82 \\
\hline\hline
\end{tabular}
\label{tab:m3}
\end{table}

\begin{table}[h!]
\centering
\caption{Model 1 (\texttt{M1})'s Performance in Big Five Traits}
\vspace{0.5cm}
\begin{tabular}{lcccccc}
\toprule
\textbf{Big Five} & \textbf{MSE} & \textbf{MAE} & \textbf{R²} & \textbf{Accuracy} & \textbf{F1 Score} \\
\midrule
Agreeableness & 0.08 & 0.25 & 0.34 & 0.59 & 0.45 \\~\\
Conscientiousness & 0.08  & 0.22 & 0.42 & 0.75 & 0.64 \\~\\
Extraversion & 0.22  & 0.35 & 0.34  & 0.60 & 0.44 \\~\\
Neuroticism & 0.07   & 0.25 & 0.33 & 0.51 & 0.34 \\~\\
\textcolor{teal}{Openness} & \textcolor{teal}{0.05}   & \textcolor{teal}{0.19} & \textcolor{teal}{0.43} & \textcolor{teal}{0.75} & \textcolor{teal}{0.66}     \\
\hline\hline
\end{tabular}
\label{tab:m1}
\end{table}

Comparing Table~\ref{tab:m3} and Table~\ref{tab:m1}, we found that M3 has similar performance to M1 in predicting Extraversion. However, in the other four dimensions—Agreeableness, Conscientiousness, Neuroticism, and Openness—M3 remarkably outperforms M1. This trend indicates that while both models are competent in predicting Extraversion, M3 has a superior capability in capturing the nuances of the other personality traits. The similar performance pattern observed between M3 and M1 in Extraversion is also evident in the other three models. This suggests that certain dimensions of personality traits, like Extraversion, might be easier to predict accurately across different models. However, the enhanced performance of M3 in the remaining four dimensions underscores the effectiveness of the additional strategies and modifications implemented in M3. This analysis highlights the importance of evaluating models across multiple dimensions to fully understand their strengths and weaknesses. While M1 provides a solid baseline, the advancements in M3 demonstrate significant improvements in the overall predictive accuracy and reliability of personality trait assessment.

In summary, while all models performed adequately, M3 demonstrated superior performance across all evaluated metrics. The detailed results for the other models, along with their performance across each personality trait dimension, are included in the Appendix for further reference and analysis.

\section{Discussion}
\label{headings}
We found that the original fine-tuning method without special strategies performed similarly to those models trained with mixed strategies. For instance, M5 employed very complex strategies. However, the model performance did not improve significantly. Some studies indicate that fine-tuning with MLP or SVM can greatly enhance model performance. Despite of this, such strategies did not yield improvements in our models trained to produce continuous values. Even when MLP and Optuna-selected optimized hyperparameters were used to train M2, the expected improvements were not observed.

Nevertheless certain strategies did result in noticeable improvements when some conditions are given. For example, the implementation of sophisticated learning rate schedulers and regularization techniques, such as dropout, led to immediate performance enhancements when MLP and the pre-training RoBERTa were provided. However, such strategies took effect when some setups were given. In M2, the same conditions and strategies were given but the results have not been significantly improved. M3 had the best performance when some parameters were adjusted. This might suggest that the critical factors for improving model performance may lie in specific advanced training techniques rather than the mere complexity of the strategies and architectures employed. It highlights the importance of identifying and applying the right \textbf{mix of techniques} to enhance the model's ability to predict continuous values accurately.

In comparison to relevant research, Table~\ref{table:combined_results} provides an overview of SOTA models evaluated on the ``essay'' dataset \citep{lin2023novel}. It is noteworthy that the ``essay'' dataset traditionally requires binary classification for each dimension. In contrast, our models are capable of generating continuous values for each dimension, demonstrating a significant advancement in the methodology for personality trait assessment. This comparison show that our models outperform the binary models.

\begin{table}[h!]
\centering
\caption{Evaluation Results of ``Essays'' Feature Selection (binary results) with Accuracy and F1 Score}
\vspace{0.1cm}
\begin{tabular}{lccccccc}
\toprule
\textbf{Algorithm} & \textbf{EXT} & \textbf{NEU} & \textbf{AGR} & \textbf{CON} & \textbf{OPN} \\
\midrule
Not select & 69.16 (59.98) & 70.43 (57.99) & 63.68 (57.93) & 71.996 (58.97) & 72.55 (59.59) \\
GA         & 72.88 (63.71) & 70.58 (63.86) & 67.15 (60.73) & 73.07 (66.198) & 73.02 (62.81) \\
GOA        & 72.86 (64.26) & 70.66 (64.44) & 66.99 (61.297) & 72.78 (65.79) & 73.13 (63.32) \\
SSA        & 73.12 (67.02) & 72.56 (65.94) & 67.87 (62.497) & 74.42 (67.38) & 77.44 (67.62) \\
MVO        & 74.79 (67.64) & 70.80 (66.55) & 71.58 (63.11) & 74.52 (67.99) & 77.18 (68.22) \\
GWO        & 75.81 (67.80) & 71.76 (66.87) & 74.46 (66.72) & 74.15 (67.91) & 74.29 (67.81) \\
IDGWOFS    & 75.97 (69.79) & 71.74 (67.89) & 73.17 (69.84) & 76.77 (70.90) & 77.74 (69.8) \\
\bottomrule
\end{tabular}
\label{table:combined_results}
\end{table}


Ultimately, our methods, employing existing models and mixed strategies, have achieved impressive results. These strategies can be applied to a variety of classification tasks and related applications \citep{sun2024automatic}.




\bibliography{reference}
\bibliographystyle{tmlr}

\appendix
\section*{Appendix}

\begin{table}[h!]
\centering
\caption{Model 2 (\texttt{M2})'s Performance in Big Five Traits}
\vspace{0.5cm}
\begin{tabular}{lcccccc}
\toprule
\textbf{Big Five} & \textbf{MSE} & \textbf{MAE} & \textbf{R²} & \textbf{Accuracy} & \textbf{F1 Score} \\
\midrule
Agreeableness & 0.08 & 0.24 & 0.34 & 0.6 & 0.45 \\~\\
Conscientiousness & 0.07  & 0.22 & 0.43 & 0.75 & 0.65    \\~\\
Extraversion & 0.22  & 0.35 & 0.35  & 0.60 & 0.44 \\~\\
Neuroticism & 0.07   & 0.23 & 0.33 & 0.57 & 0.53 \\~\\
Openness & 0.05   & 0.19 & 0.43 & 0.76 & 0.66     \\
\hline\hline
\end{tabular}
\label{tab:m1}
\end{table}

\begin{table}[h!]
\centering
\caption{Model 4 (\texttt{M4})'s Performance in Big Five Traits}
\vspace{0.5cm}
\begin{tabular}{lcccccc}
\toprule
\textbf{Big Five} & \textbf{MSE} & \textbf{MAE} & \textbf{R²} & \textbf{Accuracy} & \textbf{F1 Score} \\
\midrule
Agreeableness & 0.08 & 0.25 & 0.34 & 0.59 & 0.45 \\~\\
Conscientiousness & 0.05  & 0.2 & 0.42 & 0.75 & 0.65 \\~\\
Extraversion & 0.22  & 0.35 & 0.34  & 0.60 & 0.44 \\~\\
Neuroticism & 0.08   & 0.25 & 0.33 & 0.51 & 0.34 \\~\\
Openness & 0.05   & 0.19 & 0.43 & 0.75 & 0.65     \\
\hline\hline
\end{tabular}
\label{tab:m4}
\end{table}

\begin{table}[h!]
\centering
\caption{Model 5 (\texttt{M5})'s Performance in Big Five Traits}
\vspace{0.5cm}
\begin{tabular}{lcccccc}
\toprule
\textbf{Big Five} & \textbf{MSE} & \textbf{MAE} & \textbf{R²} & \textbf{Accuracy} & \textbf{F1 Score} \\
\midrule
Agreeableness & 0.09 & 0.25 & 0.33 & 0.6 & 0.45 \\~\\
Conscientiousness & 0.05  & 0.22 & 0.43 & 0.75 & 0.64 \\~\\
Extraversion & 0.22  & 0.35 & 0.34  & 0.60 & 0.44 \\~\\
Neuroticism & 0.08   & 0.25 & 0.34 & 0.51 & 0.34 \\~\\
Openness & 0.05   & 0.2 & 0.43 & 0.75 & 0.66     \\
\hline\hline
\end{tabular}
\label{tab:m5}
\end{table}

\end{document}